\pdfoutput=1

\documentclass[11pt]{article}
\usepackage{enumitem}
\usepackage{float}
\usepackage{fancyhdr}
\usepackage[final]{latex/acl}
\usepackage{subfigure}

\usepackage{times}
\usepackage{latexsym}
\usepackage{float}
\usepackage[T1]{fontenc}

\usepackage[utf8]{inputenc}

\usepackage{microtype}
\usepackage[table]{xcolor}

\usepackage{inconsolata}

\usepackage{graphicx}
\usepackage{multirow}
\usepackage{adjustbox}
\usepackage{colortbl}
\usepackage{amsmath}
\usepackage{graphicx}
\usepackage{booktabs}
\usepackage[ruled,vlined,linesnumbered]{algorithm2e}

\usepackage{caption}
\usepackage[ruled,vlined,linesnumbered]{algorithm2e}
\usepackage{amsmath}
\SetKwComment{Comment}{$\triangleright$\ }{}
\SetKwInput{KwInput}{Input}
\SetKwInput{KwParam}{Parameter}
\SetKwInput{KwOutput}{Output}
\SetNlSty{texttt}{}{:}
\usepackage[symbol]{footmisc}

\usepackage{subcaption}
\title{METok: Multi-Stage Event-based Token Compression \\for Efficient Long Video Understanding}
    \author{Mengyue Wang\textsuperscript{\rm 1,7} \quad  Shuo Chen\textsuperscript{\rm 2,6,7} \quad  Kristian Kersting\textsuperscript{\rm 3,4,5} \quad \\
    \textbf{Volker Tresp\textsuperscript{\rm 2,7} \quad  Yunpu Ma\textsuperscript{\rm 2,7 \footnotemark[2]}} \\
$^{1}$Technical University of Munich $\;$ $^{2}$ LMU Munich $\;$  $^{3}$DFKI SAINT $\;$ $^{4}$Hessian AI $\;$ \\
 $^{5}$TU Darmstadt $\;$ 
$^{6}$Konrad Zuse School of Excellence in Reliable AI (relAI)   \\
$^{7}$Munich Center for Machine Learning (MCML)  
}

\begin{document}
\pagestyle{fancy}
\fancyhf{}
\fancyhead[L]{Published as a conference paper at EMNLP 2025}
\fancyhead[R]{}
\maketitle
\begin{abstract}
Recent advances in Vision Large Language Models (VLLMs) have significantly enhanced their ability to understand video content. Nonetheless, processing long videos remains challenging due to high computational demands and the redundancy present in the visual data. In this work, we propose \textbf{METok}, a training-free, \textbf{M}ulti-stage \textbf{E}vent-based \textbf{Tok}en compression framework designed to accelerate VLLMs' inference while preserving accuracy. METok progressively eliminates redundant visual tokens across three critical stages: (1) event-aware compression during vision encoding, (2) hierarchical token pruning in the prefilling stage based on semantic alignment and event importance, and (3) a decoding-stage KV Cache optimization that further reduces memory consumption. Our experiments on diverse video benchmarks demonstrate that METok achieves an optimal trade-off between efficiency and accuracy by dynamically selecting informative visual tokens. For instance, equipping LongVA-7B with METok realizes an 80.6\% FLOPs reduction and 93.5\% KV Cache memory savings, all while maintaining comparable or even superior accuracy. The code is available \href{https://github.com/mnyuew/METok}{here}. 
\end{abstract}
\addtocounter{footnote}{2}
\footnotetext{\;Corresponding author: cognitive.yunpu@gmail.com}

\section{Introduction}
\label{sec:intro}
\begin{figure}[h]
  \centering
  \includegraphics[width=\linewidth]{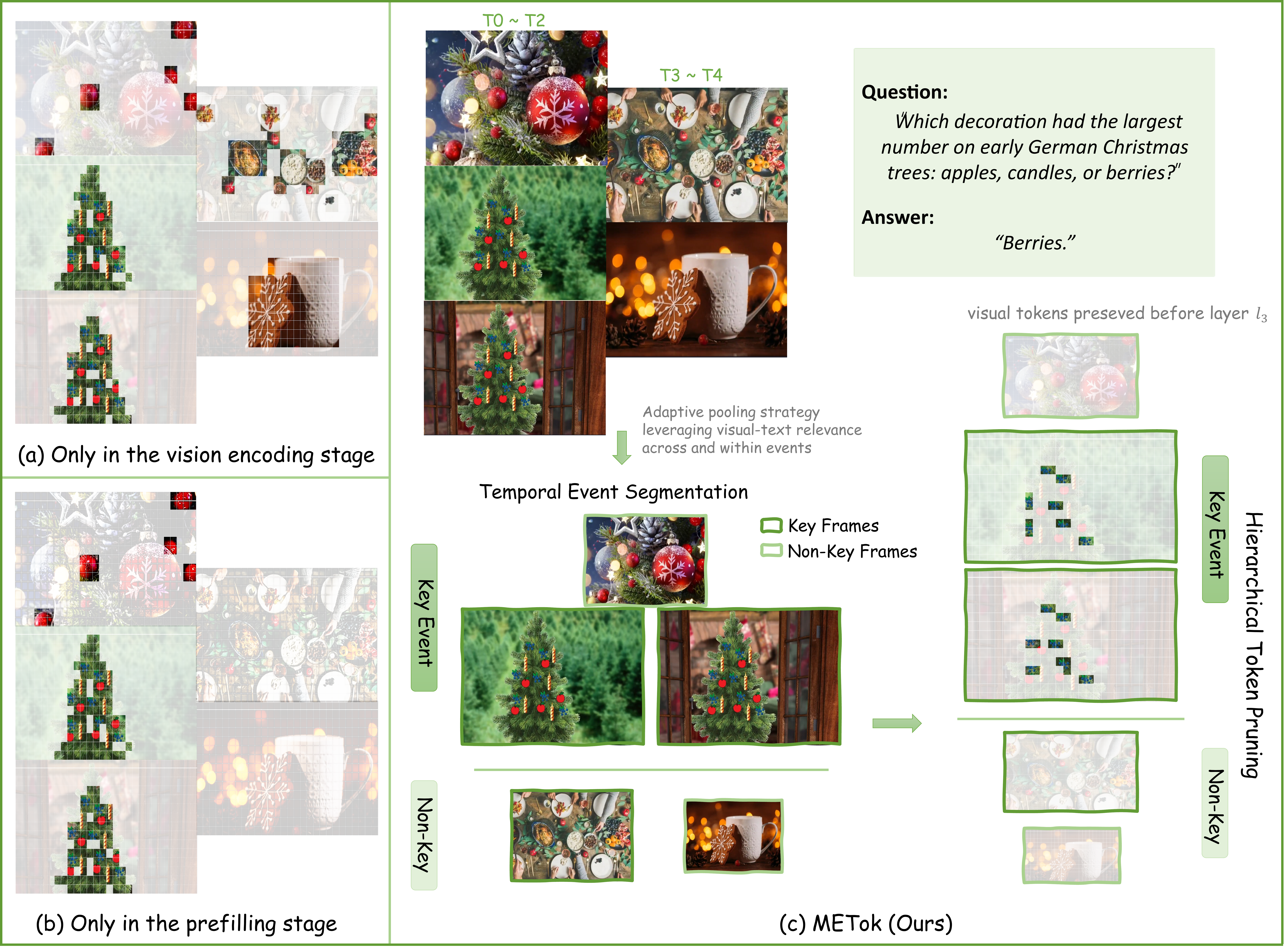}
    
   \caption{Comparison of visual token compression methods: (a) methods that only compress tokens during the vision encoding stage, such as VisionZip~\cite{visionzip}; (b) methods that drop tokens only in the prefilling stage like FastV~\cite{fastv}; (c) METok (ours), which progressively removes irrelevant tokens in three stages (vision encoding, prefilling, and decoding).}

   \label{fig:comparison}
   \vspace{-10pt}
\end{figure}
\vspace{-5pt}
Vision Large Language Models (VLLMs)~\cite{videollama2,qwen2vl,videollava,llamavid,videoinsta,llavasteering} have recently achieved remarkable success in various tasks such as video question-answering, temporal reasoning, and grounding. However, extending these models to efficiently understand long videos remains a major challenge. Current VLLMs typically encode each frame into hundreds of tokens, leading to prohibitive computational and memory overhead as sequence lengths scale. Beyond the resource burden, this visual token proliferation also introduces semantic redundancy, where repetitive or low-salience tokens obscure truly informative visual cues. As the number of frames increases, critical visual information becomes diluted, ultimately degrading both model efficiency and performance. 

To address these challenges, previous work~\cite{tome,fastv,purmerge,pllava,fu2024lazyllm,ren2023testa} has explored various token compression strategies to optimize VLLM inference. However, most of these approaches operate on a single-stage level, and do not fully account for how the relevance of visual information evolves across different layers and stages of inference. Such ignorance may lead to overlooking cross-modal alignment or discarding features that may only become salient in deeper layers. 

Consequently, an important question arises: \textit{How can we effectively reduce redundant visual tokens while preserving informative spatiotemporal content throughout the entire inference process?} 

To answer this question, we analyze existing VLLMs across inference stages and derive two key insights: (1) The critical video information is task-dependent and temporally sparse, underscoring the need to preserve spatiotemporal relationships while retaining text-relevant visual tokens. (2) Visual tokens primarily contribute in shallow LLM layers, with their influence diminishing in deeper layers as attention shifts to text tokens. 

Building upon these observations, we propose \textbf{METok}, a \textbf{M}ulti‐stage \textbf{E}vent‐based \textbf{Tok}en compression framework that integrates visual‐text alignment with hierarchical token reduction, without requiring additional training. METok operates in three stages: During the vision encoding stage, it segments videos into temporally coherent events and identifies key visual tokens while heavily compressing non-key ones. In the prefilling stage, it performs layer-wise hierarchical token pruning guided by attention and event importance. Finally, METok discards visual tokens from the KV Cache starting at the pruning boundary used in prefilling. We validate METok across diverse models and benchmarks, demonstrating that it can reduce computational cost by up to 80.6\% and KV cache memory by over 90\%, all while maintaining or even improving accuracy, and requiring no additional training. These three stages jointly improve memory and computational efficiency without compromising model performance, making METok an effective solution for long video understanding in VLLMs. The main contributions are summarized as follows:
\begin{itemize}
    \item We propose \textbf{METok}, a training-free, plug-and-play token compression framework for long video understanding, applicable across the entire VLLM inference pipeline.
    \item METok employs stage-specific compression strategies that preserve text-relevant and spatiotemporal information through semantic-aware token pruning.
    \item Extensive experiments demonstrate that METok substantially reduces computational demands while maintaining or improving performance over base VLLMs.
\end{itemize}
\vspace{-2pt}
\section{Related Work}
\subsection{Vision Large Language Models}
The rapid growth of Large Language Models (LLMs) ~\cite{qwen,vicuna,llama} has further accelerated the development across a wide range of areas~\cite{gentkg, webpilot, longva, prism} and notably in VLLMs, where strong language backbones are extended with vision encoders. To extend these capabilities from images to videos, recent works have proposed various strategies for integrating temporal visual inputs into LLMs. LLaVA-OneVision\cite{llavaov} unifies image and video tasks within a single model that supports cross-modal transfer. LongVA\cite{longva} scales video comprehension by extrapolating LLM's context length, allowing inference over thousands of visual tokens. However, most VLLMs still process video frames independently, encoding each frame into its own set of tokens. While suitable for short clips, this approach becomes prohibitively expensive for long videos due to the quadratic complexity of self-attention. 

\vspace{-4pt}
\subsection{Token Compression}
Token compression techniques~\cite{tome,aim,framefusion,visionzip,lookm,mustdrop,shen2024tempme,dycoke,pyramidDrop} in VLLMs can be categorized by the inference stage at which they operate. For instance, AuroraCap~\cite{AuroraCap} merges similar visual tokens within transformer layers, LVC~\cite{lvc} introduces a parameter-free query-attention video compression mechanism, St3~\cite{st3} prunes inattentive visual tokens progressively across LLM layers, VideoChat-Flash~\cite{videochatflash} segments video and applies hierarchical compression in two stages, and FastVid~\cite{fastvid} adopts a density-based token pruning strategy to maintain essential information. Methods such as VisionZip~\cite{visionzip} and DivPrune~\cite{alvar2025divprune} compress visual tokens early in the vision encoding stage. Prefilling-stage token compression approaches like FastV~\cite{fastv} remove redundant visual tokens after text interaction begins, typically by analyzing attention weights within LLM to drop those contributing the least to multimodal reasoning. While these techniques~\cite{aim,vlcache,streamingllm} bring efficiency gains, they often operate in isolation, overlooking the cumulative impact of redundancy across stages. In contrast, METok integrates token compression across all stages with spatiotemporal awareness, enabling efficient and effective processing of long video sequences. 

\vspace{-2pt}
\section{Method}
\begin{figure*}[h]
  \centering
  \includegraphics[width=1\linewidth]{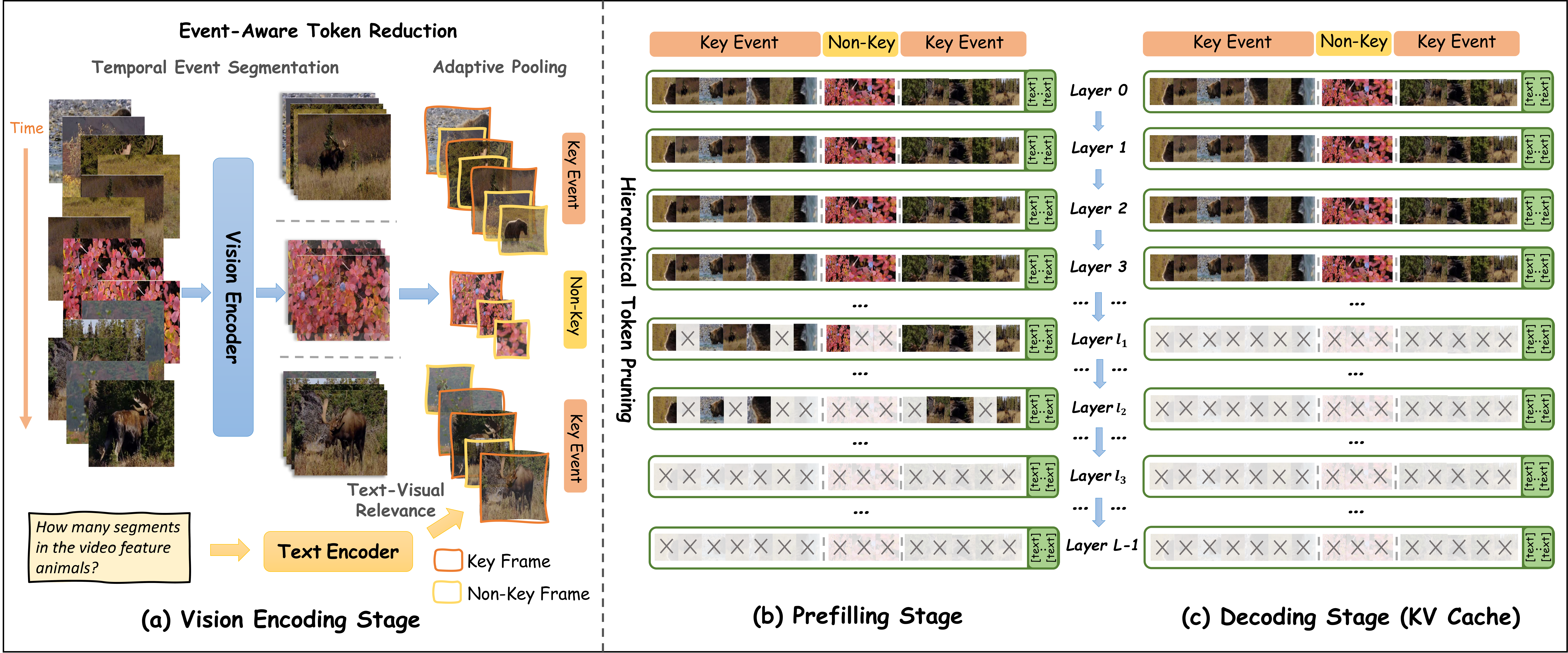}
    
   \caption{The architecture of METok. In the vision encoding stage, METok first segments video into events and compresses key events and frames based on visual-text relevance. Then, it hierarchically prunes redundant visual tokens using attention and event importance in the prefilling stage. Finally, the prefill-driven KV Cache Optimization further removes visual tokens starting at layer $l_1$.}
   \label{fig:METok}
   \vspace{-5pt}
\end{figure*}

\subsection{Preliminary}
\noindent \textbf{Task Formulation.} Given a video $V$ and a text
input $c$, the goal is to generate a textual response $y$ by maximizing the conditional probability 
\vspace{-2pt}
\begin{equation}
    \max_{y}\; p \left(y \mid V,\,c)\right.
\end{equation}

\noindent \textbf{Vision encoding stage.} Each frame is divided into $N$ patches $\{\mathbf{x}_i\}_{i=1}^N$, projected into embeddings with positional encoding and processed through $L$ layers of Multi-Head Self-Attention (MHSA)~\cite{transformer}:
\begin{equation}
  \text{Attention}(\mathbf{Q}, \mathbf{K}, \mathbf{V}) = \text{softmax}\left(\frac{\mathbf{Q}\mathbf{K}^\top}{\sqrt{d/H}}\right)\mathbf{V}, 
\end{equation}
\begin{figure}[H]
  \centering
  \includegraphics[width=\linewidth]{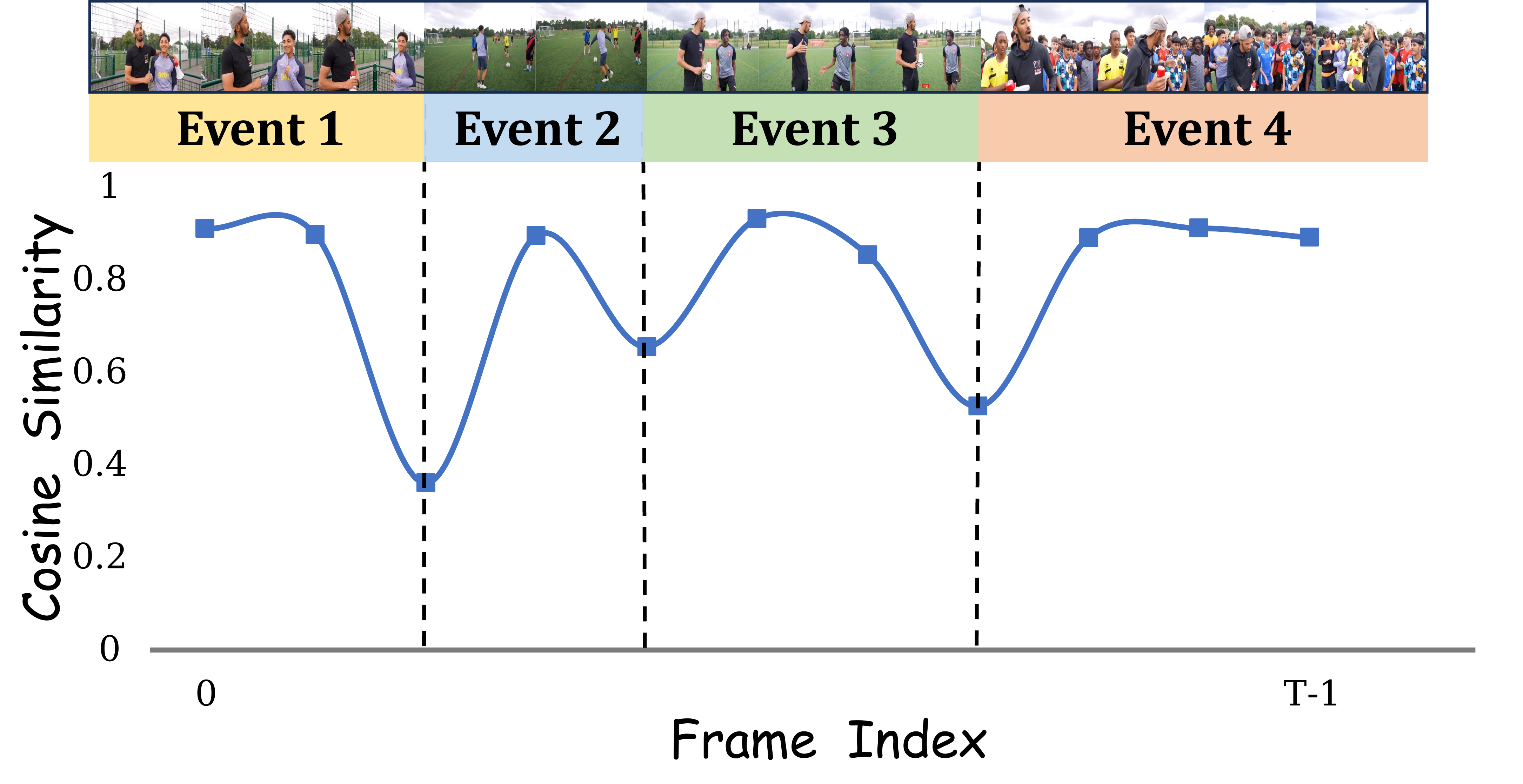}
    
   \caption{The visualization of temporal event segmentation based on the similarity of adjacent frames.}
   \label{fig:Event}
\end{figure}\vspace{-8pt}
where $\mathbf{Q}, \mathbf{K}, \mathbf{V} \in \mathbb{R}^{N \times d/H}$ are query, key, and value matrices distributed across $H$ heads, and $d$ denotes the feature dimension of each token.

\noindent \textbf{Prefilling stage.}
 The model encodes the concatenated input $\mathbf{X} = [\mathbf{X}_{\text{vis}}; \mathbf{X}_{\text{text}}] \in \mathbb{R}^{T \times d}$, and caches key-value pairs $\{\mathbf{K}_l, \mathbf{V}_l\}$  for each transformer layer $l$ using projection matrices $\mathbf{W}_l^K$ and $\mathbf{W}_l^V$ as follows:
\vspace{-2pt}
\begin{equation}
    \mathbf{K}_l = \mathbf{X}\mathbf{W}_l^K, \quad \mathbf{V}_l = \mathbf{X}\mathbf{W}_l^V. 
\end{equation}

\noindent \textbf{Decoding stage.}
The decoding process follows an autoregressive pattern, in which each token is predicted sequentially by referencing previously generated tokens and stored key‐value (KV) pairs.


\subsection{METok}
METok follows a structured three‐stage approach tailored for long‐video inference to progressively remove redundant visual tokens, achieving an optimal balance between computational efficiency and preservation of spatiotemporal semantics throughout the entire pipeline, as shown in Figure~\ref{fig:METok}.

\vspace{-4pt}
\subsubsection{Vision Encoding: Event-Aware Token Reduction}
\textbf{Temporal Event Segmentation.} 
Long videos are naturally composed of multiple semantically distinct events that unfold over time. When all frames are treated equally, this temporal structure is ignored, resulting in inefficient processing and a diluted representation of meaningful visual content. To better reflect this structure, METok segments the video into meaningful temporal events based on content dynamics. Some existing methods like Chat-Univi~\cite{chatunivi} with DPC-KNN~\cite{dpcknn}, use clustering to capture event structure, but often ignore temporal continuity and produce fragmented segments. In contrast, METok detects meaningful transitions between frames to identify coherent event boundaries.

Given frame-level visual embeddings $ V = \{v_i\}_{i=1}^{T} \in \mathbb{R}^{T \times N \times d} $, where $T$ is the number of frames, $N$ is the number of tokens per frame, and $d$ is the embedding dimension, the cosine similarity between adjacent frame embeddings $v_i$ and $v_{i+1}$ is computed as 
\begin{equation}
    S_i = \frac{v_i \cdot v_{i+1}}{\| v_i \| \| v_{i+1}\|}, \quad i\in [0, T-1]. 
\end{equation}

As shown in Figure~\ref{fig:Event}, a lower similarity score indicates a significant change in content, suggesting a potential event boundary. We select the $k{-}1$ lowest $S_i$ scores as event boundaries, dividing the video into $k$ events. Let $E$ denote the set of events.



\noindent \textbf{Key Visual-Text Sematic Identification.}  After event segmentation, METok estimates the text relevance of each frame to distinguish semantically important content from less informative visual information. we observed that many existing VLLMs, such as LLaVA-OneVision~\cite{llavaov} and LongVA~\cite{longva}, employ CLIP~\cite{clip} or SigLIP~\cite{siglip} ViT-L as their vision encoders, which are pretrained for visual-text alignment. To leverage this, METok reintroduces the associated text encoder to compute the cross-modal similarity score $S_{v_i,t}$ between the visual embeddings $v_i \in \mathbb{R}^{ N \times d} $ of each frame and the encoded text embedding $t \in \mathbb{R}^{1 \times d}$: 
\vspace{-2pt}
\begin{equation}
    S_{v_i,t} = \frac{ v_i \cdot t}{ \| v_i \|\| t \|}, \quad i\in [0, T-1]. 
\end{equation}
We average these scores per event and rank all $k$ events accordingly. The top $\lceil \alpha \cdot k \rceil$ events ($0 < \alpha < 1$) are designated as key events $E_{\text{key}}$ while the remaining are considered non-key events $E_{\text{non-key}}$. Within each event, whether key or non‐key, METok further selects the top $\beta$ proportion of frames ($0 < \beta < 1$) that convey richer textual semantics as key frames, treating all remaining frames as non‐key frames.


\noindent \textbf{Adaptive Pooling Strategy.} METok applies an adaptive pooling strategy to refine compression while preserving temporal relationships. It applies \textit{different pooling granularities to key and non-key frames of key and non-key events}. Specifically, we define different pooling strides $s_1$ and $s_2$ for key and non-key frames in each event of $E_{\text{key}}$. To maintain key events at a higher resolution while aggressively downsampling non-key events, we further adjust the pooling strides $s_1$ and $s_2$ by a factor of $\tfrac{1}{\alpha} > 1$ for events in $E_{\text{non-key}}$, ensuring that non-key events undergo higher pooling. This approach ensures the retention of critical event and frame details while removing superfluous content, thereby optimizing both event-level and frame-level token compression for long video understanding.

\subsubsection{Prefilling: Hierarchical Token Pruning}
Although the vision encoding stage filters out many redundant tokens, some visually marginal content still remains. If not further processed, these tokens persist through the model's inference, occupying memory and compute without meaningfully contributing to the final output. To address this, METok introduces a hierarchical token pruning strategy in the prefilling stage, ensuring a progressive refinement of visual token selection. 

The design builds on the observation that early layers in LLMs primarily extract low-level visual features while deeper layers gradually emphasize semantically relevant elements. Leveraging this shift, METok prunes visual tokens in a layer-wise manner guided by visual-text attention scores. At each layer, only the top-ranked tokens—those most semantically aligned with the text input—are retained, ensuring that models allocate resources to the informative content throughout inference. 

Concretely, METok retains a higher proportion of tokens for key events and prunes more aggressively for non-key events, with the relative pruning rates modulated by the key event ratio $\alpha$ (see Eq.~\ref{eq:unkey}) The pruning process unfolds across three compression levels defined by the layer boundaries $L_H = [l_1, l_2, l_3]$. Token retention ratios are expressed relative to the number of visual tokens originally passed into the LLM. The visual token retention ratio for key events, $R_{\text{key}}(l)$, and non-key events, $R_{\text{non-key}}(l)$, across different layers $l$ are defined as:   
\begin{equation}
    R_{key}(l) = 
    \begin{cases} 
    1, & \text{if } l < l_1 \\
    r, & \text{if } l_1 \leq l < l_2 \\
    r^2, & \text{if } l_2 \leq l < l_3 \\
    0, & \text{if } l > l_3
    \end{cases}
    \label{eq:key_ratio_r}
\end{equation}
\begin{equation}
    R_{non\_key}(l) = 
        \begin{cases} 
        1, & \text{if } l < l_1 \\
        \alpha \cdot r, & \text{if } l_1 \leq l < l_2 \\
        0, & \text{if } l > l_2
        \end{cases}
    \label{eq:unkey}
\end{equation}
where $r$ is a pruning factor ($0 < r < 1$), and $\alpha$ is the proportion of key events. 

This progressive pruning process ensures that early layers preserve necessary visual information, intermediate layers refine essential semantic information, and later layers retain only the most relevant text-aligned features, allowing for efficient and effective processing of long videos.

\subsection{Decoding: Prefilling-driven Optimization}

Despite the effectiveness of prior token pruning in reducing visual redundancy, optimizing memory efficiency throughout the entire inference process-particularly during decoding-remains challenging. Most VLLMs still maintain all visual tokens in the KV Cache across all layers, resulting in substantial memory overhead. 

\begin{figure}[h]
  \centering  \includegraphics[width=\linewidth]{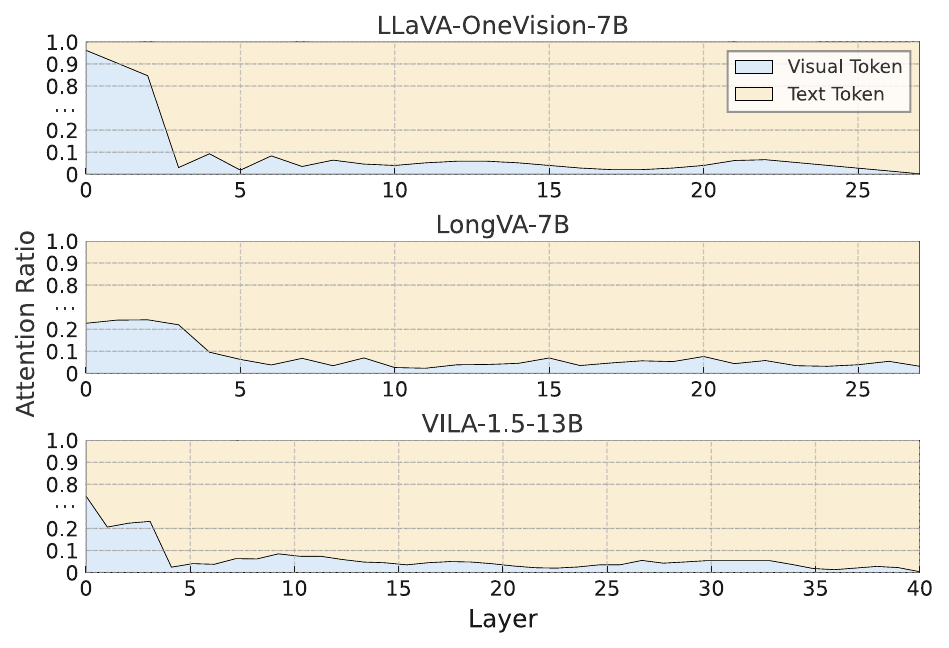}
   \caption{Layer-wise attention ratio between visual and text tokens during decoding in various VLLMs.}
   \label{fig:attention}
\end{figure}

To better understand this inefficiency, we analyze the attention distribution between visual and text tokens across decoding layers in various VLLMs like VILA-1.5~\cite{lin2024vila}, as shown in \ref{fig:attention}. Notably, attention to visual tokens drops sharply after the early decoding layers, suggesting that these tokens contribute little to response generation in deeper layers. In contrast, text tokens consistently receive high attention across all layers, underscoring their continued relevance. 
Motivated by this, METok refines the KV Cache by aligning it with the hierarchical pruning strategy from the prefilling stage. Specifically, while both visual and text tokens are cached in the early decoding layers, METok removes entire visual tokens from the KV Cache starting at layer $l_1$—the same boundary where pruning begins during prefilling. This selective preservation substantially reduces memory consumption while maintaining essential multimodal context for accurate generation.
\section{Experiments}
\begin{table*}[h]
    \centering
\begin{adjustbox}{width=\linewidth,center}
\renewcommand{\arraystretch}{1.1}
\setlength{\tabcolsep}{1.5mm}
\begin{tabular}{lcccccccccccccc}
\toprule  \multirow{2}{*}{\textbf{Model}}  & \multirow{2}{*}{\centering \textbf{\#Frames}} &\multicolumn{1}{c} {\textbf{FLOPs}} & \multirow{2}{*} {\textbf{MVBench}} & \multirow{2}{*} {\textbf{Egoschema} }  & \multirow{2}{*} {\textbf{MLVU}} & \multicolumn{2}{p{4cm}}{\centering \textbf{VideoMME (\textit{w/o sub.})} } & \multicolumn{2}{p{4cm}}{\centering \textbf{VideoMME (\textit{w sub.})} }   \\ \cline{7-8} \cline{9-10} &
& (TB) &&&& Overall & Long & Overall & Long  \\
\rowcolor{gray!10} Duration & & & 16 sec& 179.8 sec & 3$\sim$120 min & 1$\sim$60 min & 30$\sim$60 min & 1$\sim$60 min & 30$\sim$60 min\\

\midrule

LLaVA-OneVision-0.5B~\citep{llavaov} & 32 & 3.9  & 45.5 & 26.8  &  50.3 & 44.0 & 35.4& 43.5 & 38.7 \\
FastV~\cite{fastv} & 32 &2.2 &\textbf{45.0} & \underline{26.5} & \underline{50.8} & 42.4& 35.8 & 46.4 & 39.4 \\
VisionZip~\citep{visionzip} & 32 & 1.7 & 41.4 & \textbf{27.3} & 44.4 & 37.6 & 35.6 & 41.9 & 36.1 \\
DivPrune~\cite{alvar2025divprune} & 32 & \textbf{1.2} & 44.2 & 26.1 & 37.4 &\underline{42.6} & \underline{36.0} & \underline{46.4} & 
\underline{39.5}\\
\rowcolor{YellowGreen!20} METok (Ours) & 32 & \underline{1.3}  & \underline{44.7} & 25.8 & \textbf{52.1} & \textbf{42.7}& \textbf{36.4} & \textbf{47.0} & \textbf{39.7}  \\

\midrule

LLaVA-OneVision-7B~\citep{llavaov} &  32 &  71.4  & 56.7 & 60.1 & 64.7 & 58.2& 48.5&61.2 &51.4 \\
FastV~\cite{fastv} &  32 & 39.0 & \underline{57.2} & \textbf{60.1} & \underline{64.0} & \underline{58.2}& \underline{48.4}&\textbf{61.8} &\underline{51.6} \\
VisionZip~\citep{visionzip} &  32 & 27.7  & 49.4 & 56.4 & 55.0 & 49.6 & 44.6 & 55.2 & 48.4 \\
DivPrune~\cite{alvar2025divprune}  &  32 & \underline{20.9}  & 55.9 & 58.2 & 60.8 & 54.4 & 46.6 & 58.9 & 50.8 \\
\rowcolor{YellowGreen!20} METok (Ours) & 32 & \textbf{19.8} & \textbf{57.3} & \underline{59.8} & \textbf{64.7} & \textbf{58.4}& \textbf{49.8} & \underline{61.7} & \textbf{52.8}  \\

\midrule
LongVA-7B~\citep{longva}  & 128 & 241.6  & 50.8 & 43.5 & 59.0 & 52.6& 46.2&54.3 &47.6 \\
FastV~\cite{fastv}  & 128 & 133.6 & \textbf{50.9} & \underline{44.2}  & \underline{59.2} & \textbf{52.9} & 45.6 & 55.7 & 47.0 \\
VisionZip~\citep{visionzip}  & 128 & 107.9 & 46.3 & 36.2 & 52.0 & 45.4 & 37.9 & 51.4 & 41.0 \\
DivPrune~\cite{alvar2025divprune}  &  128 & \underline{50.1}  & 50.6 & 43.9 & 58.7 & \underline{52.4} & \underline{45.9} & \underline{55.8} & \underline{47.4} \\
\rowcolor{YellowGreen!20} METok (Ours) & 128 & \textbf{46.8} & \underline{50.7} & \textbf{44.4} & \textbf{60.4} & \underline{52.4} & \textbf{46.6} & \textbf{56.0} & \textbf{47.7}  \\

\midrule
VILA-1.5-13B~\citep{lin2024vila}  & 16 & 84.1 & 50.9 & 50.4 & 50.4 & 49.5 & 42.4 & 53.3 & 46.7 \\
FastV~\cite{fastv}  & 16 & 49.8 & 50.4 & 50.0 & 49.5 & 48.4 & 41.6 & 53.1 & 45.9 \\
VisionZip~\citep{visionzip}  & 16 & 29.1 & 45.8 & 47.6 & 47.1 & 43.8 & 39.4 &49.4 & 44.9 &  \\
DivPrune~\cite{alvar2025divprune}  &  16 & \underline{20.3}  & \textbf{50.7} & \underline{50.4} & \underline{50.3} & \underline{48.9} & \underline{41.8} & \underline{53.5} & \underline{46.3} \\
\rowcolor{YellowGreen!20} METok (Ours) &  16 & 
\textbf{19.6}  & \underline{50.5} & \textbf{50.5} & \textbf{50.5} & \textbf{49.5} & \textbf{42.7} & \textbf{53.9} & \textbf{48.1} \\

\bottomrule
\end{tabular}
\end{adjustbox}
\caption{Performance and efficiency comparison across different methods and benchmarks. The best result of token compression methods is \textbf{bolded} and the second best is \underline{underlined}.}
\label{tab:main}

\end{table*}
\begin{table}[!htbp]

    \centering
    \footnotesize
    \renewcommand{\arraystretch}{1.1}
    \setlength{\tabcolsep}{1.5mm}
    \begin{adjustbox}{width=\linewidth,center}
    \begin{tabular}{lcccc}
        \toprule
        \multirow{2}{*}{\textbf{Method}} & \textbf{FLOPs } & \textbf{Prefill Time  } & \textbf{KV Cache} & \multirow{2}{*}{\textbf{MLVU}} \\
        & \textbf{(TB)} &\textbf{(ms)} &\textbf{(MB)} &\\
        \midrule
        LongVA-7B & 241.6  & 1535.9  &  1012.3 & 59.0  \\
        w/FastV & 133.6  & 1004.5  & 535.1 & 59.2 \\
        w/Visionzip & 107.9  &  629.0 & 452.2 & 52.0 \\
        w/DivPrune & 50.1 & 417.7 & 240.6 & 58.7 \\
        
        \rowcolor{YellowGreen!20} w/METok (Ours) & \textbf{ 46.8 } & \textbf{378.6}  & \textbf{65.0} & \textbf{60.5} \\
        \midrule
        LLaVA-OneVision-7B & 71.4  & 400.4 & 299.8 & 64.7  \\
        w/FastV & 39.0 & 231.3 & 162.7 & 64.0 \\
        w/Visionzip & 27.7  & 155.6 & 116.2 & 55.0 \\
        w/DivPrune & 20.9 & 137.5 & 65.8 & 60.8 \\
        \rowcolor{YellowGreen!20} w/METok (Ours) & \textbf{19.8} & \textbf{131.8}  & \textbf{22.3} & \textbf{64.7} \\
        
        \bottomrule
    \end{tabular}
\end{adjustbox}
\caption{Efficiency comparison of FLOPs, Prefill Time and KV Cache Memory for token compression methods.}
\label{tab:efficiency_comparison}
\vspace{-10pt}
\end{table}


\noindent \textbf{Benchmarks.} We conduct evaluation on widely used video understanding benchmarks, including Egoschema~\cite{egoschema}, MVBench~\cite{mvbench}, MLVU~\cite{zhou2024mlvu}, and VideoMME~\cite{videomme}. Notably, VideoMME (1 min $\sim$ 1 hr) and MLVU (3 mins $\sim$ 2 hrs) target long video scenarios, making them suitable for assessing long-context comprehension. We use \texttt{lmms-eval}~\cite{zhang2024lmmseval} as our primary evaluation framework to uniformly assess performance across these diverse benchmarks. The details of these benchmarks are shown in Appendix\ref{sec:dataset}.

\subsection{Implementation Details} 

We evaluate METok using various open-source VLLMs spanning small to large scales, including LLaVA-OneVision-(0.5B, 7B)~\cite{llavaov}, LongVA-7B~\cite{longva} and VILA-1.5-13B~\cite{lin2024vila}, respectively sampling $32$, $128$, and $16$ frames. LLaVA-OneVision and VILA-1.5 adopt a SigLIP-pretrained ViT-L as their vision tower, while LongVA employs a CLIP-pretrained ViT-L. To assess computational efficiency, we measure the total FLOPs of both the prefilling and decoding stages using calflops~\cite{calflops}. Prefill Time refers to the latency required to generate the first token. More implementation details are included in the Appendix~\ref{sec:implementation}. 

\subsection{Main Results}

\noindent\textbf{Comparison with base models.} 
Table~\ref{tab:main} presents the performance of METok on LLaVA-OneVision-(0.5B, 7B), LongVA-7B and VILA-1.5-13B across representative benchmarks. METok consistently achieves substantial reductions in FLOPs while maintaining or improving accuracy, demonstrating a trade-off between efficiency and performance. In particular, METok shows clear advantages in long-context scenarios. On the long subset of VideoMME, it consistently surpasses all baselines, indicating its ability to retain essential temporal and semantic visual tokens under extended durations. The results also highlight METok’s scalability across model sizes, from LLaVA-OneVision-0.5B to VILA-1.5-13B, achieving up to 80.6\% FLOPs reduction without degrading performance. Notably, for each base model, we use a single set of hyperparameters across all benchmarks without task-specific tuning. METok consistently delivers strong performance under this unified setting, demonstrating robustness to hyperparameter variation.


\noindent\textbf{Comparison with baseline methods.} 
To further highlight METok’s effectiveness, we compare it against other training-free token compression baselines such as FastV~\cite{fastv}, VisionZip~\cite{visionzip} and DivPrune~\cite{alvar2025divprune} (Table~\ref{tab:main}). While FastV can reduce some computational overhead, relying solely on prefilling-stage compression places a notable burden on the shallow layers of the LLM, limiting FLOPs reduction. For instance, FastV reduces FLOPs by only 44.7\% on LongVA-7B, largely because its early removal of visual tokens can prematurely discard important information. VisionZip and DivPrune compress tokens during the vision encoding stage by merging or selecting based on redundancy. While they achieve higher FLOPs reduction (e.g., 63.7\%), their one-stage design often removes fine-grained details essential for temporal reasoning, leading to accuracy drops across tasks. In contrast, METok distributes compression across all three inference stages, preserving visual-text alignment and progressively pruning irrelevant tokens. This design enables it to maintain key visual-text interactions and systematically discards irrelevant tokens with minimal accuracy degradation across evaluated benchmarks. 


\noindent \textbf{Efficiency Analysis.}
METok offers substantial efficiency and memory savings while preserving strong video understanding performance. We compare FLOPs, prefill time, and KV Cache memory usage against vanilla LongVA-7B and LLaVA-OneVision-7B, as well as the single-stage methods FastV and VisionZip. As shown in Table~\ref{tab:efficiency_comparison}, METok consistently reduces FLOPs and prefill time with minimal or zero accuracy loss. For instance, with LLaVA-OneVision-7B, METok lowers FLOPs by 72.3\%, prefill time by 67.1\%, and KV Cache memory by 92.6\%, yet still matches the original MLVU accuracy. In contrast, one-stage approaches that either prune tokens prematurely (VisionZip and DivPrune) or rely solely on shallow cross-modal attention (FastV) often experience larger performance drops at comparable compression levels. By progressively pruning in the vision encoder, prefilling, and decoding stages, METok more accurately discards low-impact tokens while retaining those essential for video understanding. 


\subsection{Ablation Study}
\textbf{Multi-Stage Strategy.} 
We conduct an ablation experiment to assess the contribution of each stage. As shown in Table~\ref{tab:three-stage}, every stage reduces computational overhead while preserving or even slightly improving performance. In the vision encoding stage, our event-aware token compression strategy segments videos into meaningful events and adaptively retains visual tokens based on semantic relevance. This step alone cuts FLOPs and KV Cache usage by 42.6\% and slightly boosts accuracy, showing that early-stage filtering removes redundancy while keeping key information.

In the prefilling stage, we introduce a hierarchical token pruning mechanism guided by text-visual attention scores and event importance, dynamically dropping low-relevance visual tokens. This approach further reduces FLOPs and KV Cache by 72.3\% without sacrificing accuracy. Lastly, in the decoding stage, METok refines KV Cache retention by discarding tokens from shallower layers that no longer contribute to final outputs, resulting in an additional 92.5\% cut in KV Cache usage. These tailored strategies at each stage precisely eliminate invalid visual tokens while retaining content critical for downstream tasks.

\renewcommand{\arraystretch}{0.7}
\begin{table}[!t]
  
  \renewcommand{\arraystretch}{1}
  
  \centering
  \small
  \begin{adjustbox}{width=\linewidth,center}

  \begin{tabular}{lccc}
    \toprule
    \multirow{2}{*}{\textbf{Method}} & {\textbf{FLOPs}} &  \hspace{2mm}\textbf{KV Cache} & \multirow{2}{*}{\textbf{MLVU} } \\
     & \textbf{(TB)} & \textbf{(MB)} &   \\
    \midrule
    LLaVA-OneVision-7B & 71.4 &  299.2 & 64.7\\
    \midrule
     
     \hspace{3mm}\textbf{w/ Vision Encoding} & \multirow{2}{*}{41.1} & \multirow{2}{*}{171.8 {\color[HTML]{228B22}($\downarrow 42.6\%$)}} & \multirow{2}{*}{\textbf{64.8}}  \\
     Event-Aware Token Reduction  \\
      \hline
    \hspace{8mm}\textbf{w/ Prefilling}  & \multirow{2}{*}{19.8} &  \multirow{2}{*}{82.8 {\color[HTML]{228B22}($\downarrow 72.3\%$)}} & \multirow{2}{*}{64.7}  \\
    Hierarchical Token Pruning  \\
      \hline
    \hspace{8mm}\textbf{w/ Decoding} & \multirow{2}{*}{\textbf{19.8}} & \multirow{2}{*}{\textbf{22.3} {\color[HTML]{228B22}($\downarrow 92.5\%$)}} & \multirow{2}{*}{64.7}  \\
    Prefilling-driven Optimization  \\
    
    \bottomrule
  \end{tabular}
  
  \end{adjustbox}
  \caption{Ablation study of specific strategy at vision encoding, prefilling, and decoding stage, respectively.}
  \label{tab:three-stage}
  \vspace{-10pt}
\end{table}

\begin{figure}[h]
  \centering
  \includegraphics[width=\linewidth]{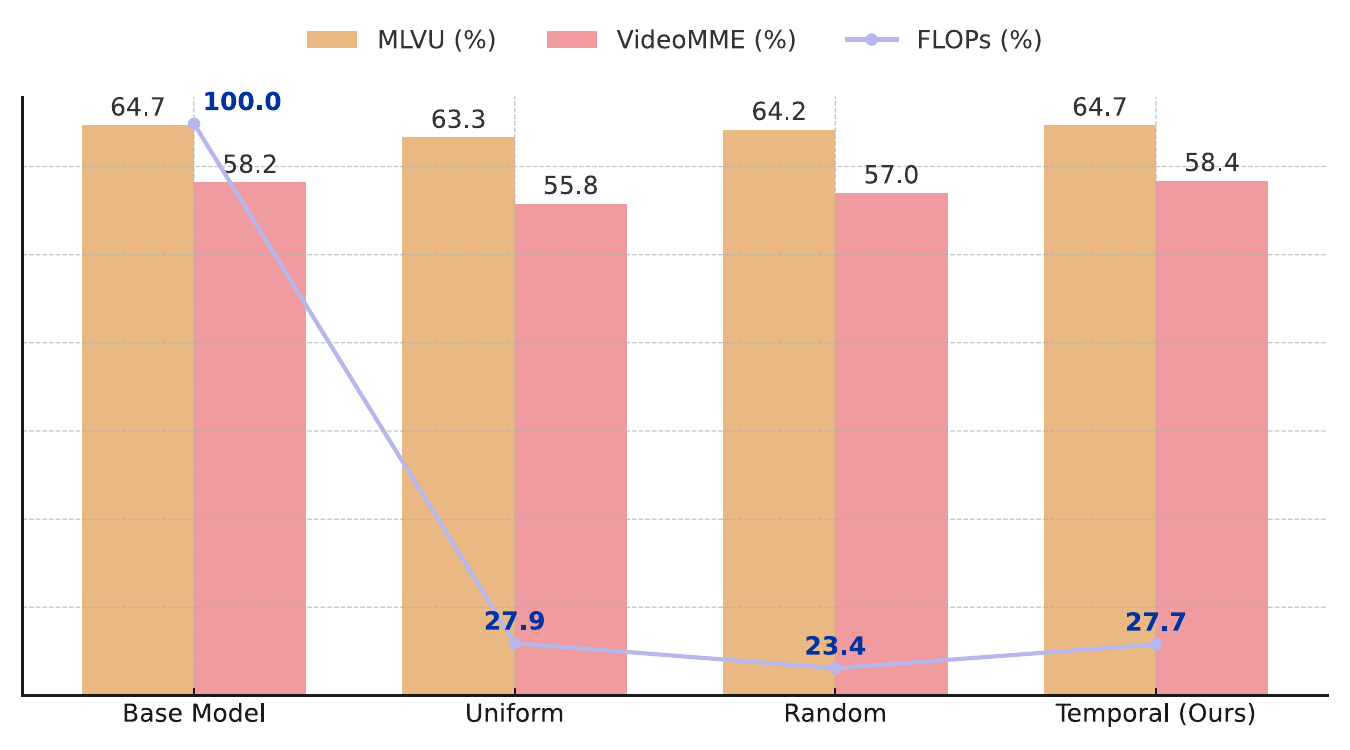}
   \caption{Comparison of FLOPs and accuracy (on MLVU and VideoMME) under different event segmentation strategies using LLaVA-OneVision-7B. }
   \label{fig:ablation_event}
\end{figure}

\noindent\textbf{Temporal Event Segmentation.}
We also compare various event segmentation strategies on LLaVA-OneVision-7B. As shown in Figure  \ref{fig:ablation_event}, uniform segmentation lowers FLOPs to 27.9\% of the baseline but noticeably hurts performance, while random segmentation further reduces FLOPs to 21.1\% yet also degrades accuracy. In contrast, our temporal segmentation strategy strikes the optimal balance by reducing FLOPs to 27.7\% while outperforming the other methods on both MLVU and VideoMME benchmarks. 

\noindent\textbf{Key Visual-Text Semantic Identification.} 
We further investigate the impact of key visual-text semantic identification at the vision encoding stage. As shown in Table~\ref{tab:ablation_key}, randomly designating key events or frames reduces FLOPs but results in unstable performance; choosing both events and frames at random yields even lower costs but significantly degrades accuracy. These drops highlight the importance of \textit{structured, semantic-aware selection}. In contrast, METok identifies semantically key events and frames, retaining only the most relevant tokens. 
\vspace{4pt}

\begin{table}[!htbp]
    
    \centering
    \begin{adjustbox}{width=\linewidth,center}
    \renewcommand{\arraystretch}{1}
    \setlength{\tabcolsep}{1.5mm}
    \begin{tabular}{lccc}
        \toprule
        \multirow{1}{*}{\textbf{Method}} & \textbf{FLOPs} &  \textbf{MLVU} & \textbf{VideoMME} \\
        
        \midrule
        Base Model & 100\% & 64.7 & 58.2 \\
        Random key events \& frames & \textbf{21.1\%} &   60.9 & 54.6   \\
        Random key frames & 27.8\% &   63.4 & 57.5   \\
        Random key events & 22.3\%  & 61.2 & 54.7 \\
        \rowcolor{YellowGreen!20} METok &  27.7\% &  \textbf{64.7}&  \textbf{58.4} \\
        \bottomrule
        \end{tabular}
    \end{adjustbox}
    \caption{Ablation of different key visual-text semantic identification strategy on LLaVA-OneVision-7B.} 
    \label{tab:ablation_key}
\end{table}
\begin{figure*}[h]
  \centering
  \includegraphics[width=\linewidth]{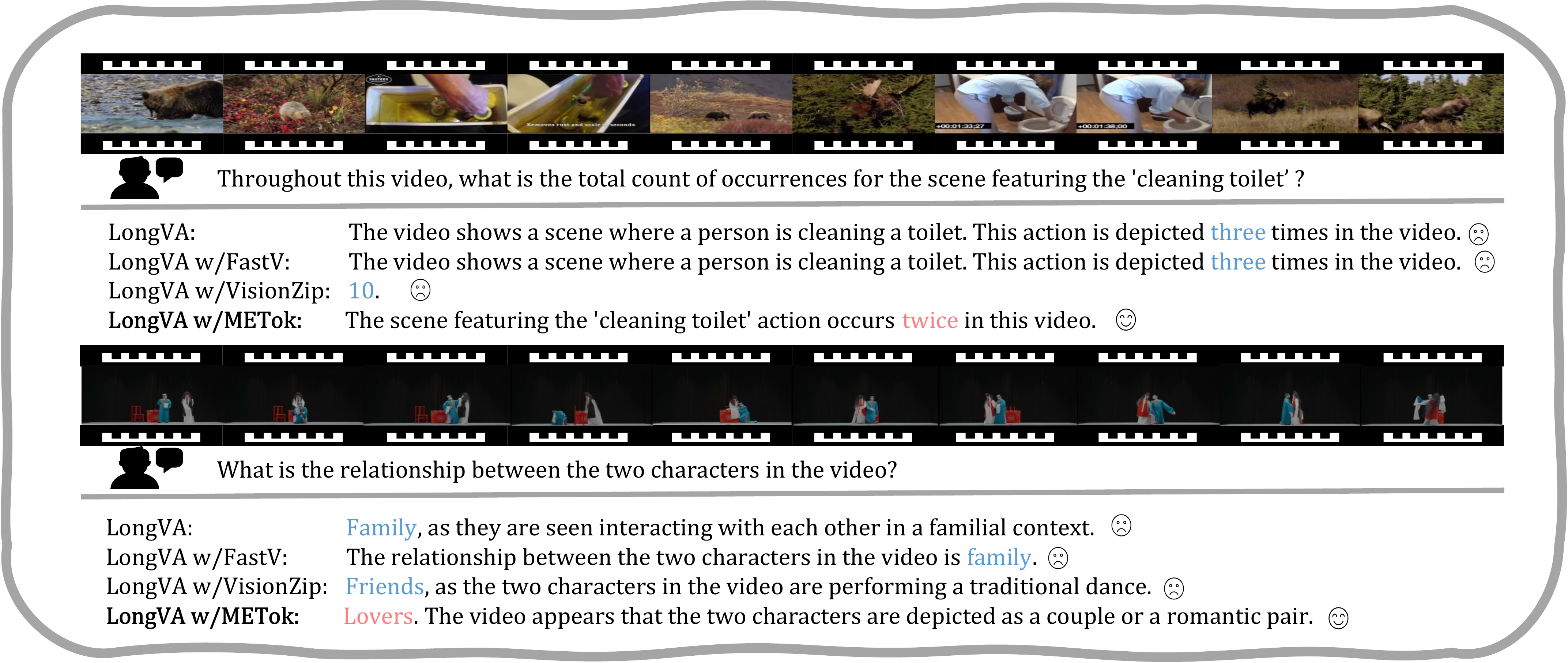}
    
   \caption{Qualitative results of METok compared to FastV and VisionZip with LongVA-7B on examples from VideoMME and MLVU benchmarks. Ground truth answers are referenced from the original benchmark annotations. }
   \label{fig:understanding}
\end{figure*} 
\begin{figure}[h]
  \centering
  \includegraphics[width=\linewidth]{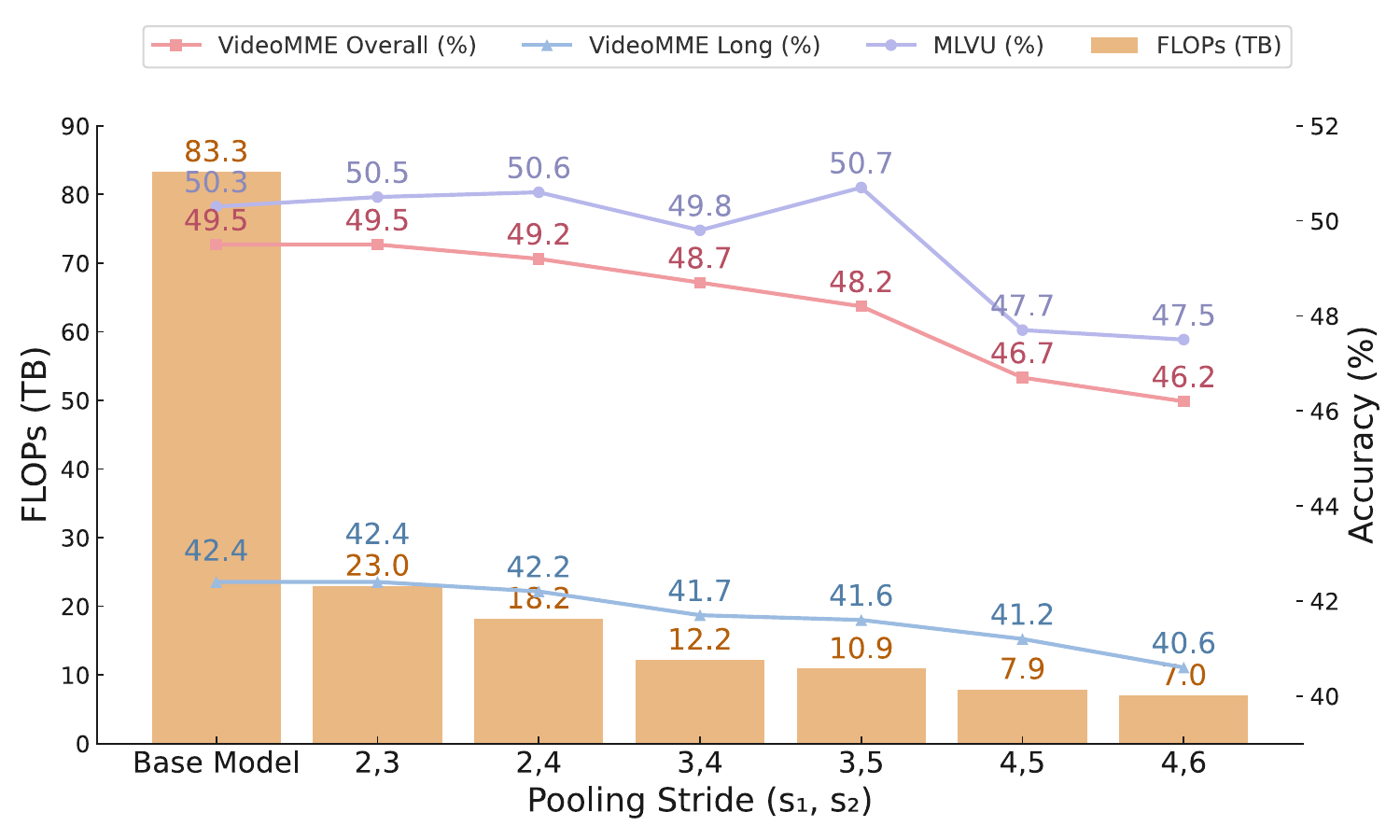}
   \caption{Ablation of adaptive pooling strategy on VILA-1.5-13B. Base Model refers to the uniform pooling setting with stride 2 for all frames.}
   \label{fig:adaptive_pooling}
   \vspace{-4pt}
\end{figure}

\noindent\textbf{Adaptive Pooling Strategy.} We ablate the adaptive pooling strategy by varying the pooling strides($s_1,s_2$), with the key event ratio $\alpha=0.5$ fixed. As shown in Figure~\ref{fig:adaptive_pooling}, smaller strides retain more visual detail and improve performance, while larger strides result in lower FLOPs at the cost of degraded accuracy. The consistent trade-off across settings confirms that METok's adaptive pooling strategy effectively balances efficiency and semantic preservation, validating the overall effectiveness of the proposed design. Notably, even under highly aggressive pooling settings that reduce FLOPs to less than 10\% of the base model, METok retains reasonable performance, highlighting its stability to hyperparameter choices.

\vspace{4pt}
\noindent\textbf{Hierarchical Token Pruning.} 
An ablation study on hierarchical token pruning with LLaVA-OneVision-7B (Table~\ref{tab:ablation_layers}) demonstrates the importance of aligning pruning with the LLM’s evolving semantic focus. Shallow pruning, which removes tokens aggressively in early layers, yields greater FLOPs savings but severely degrades accuracy by discarding critical low-level features too soon. Deep pruning, performed in later layers, better preserves essential information but reduces fewer FLOPs. In contrast, a balanced strategy achieves the best trade-off, cutting FLOPs to 27.7\% while maintaining the baseline MLVU score of 64.7. By progressively pruning across all layers and dynamically adjusting token retention based on text-visual attention scores, METok prevents premature or excessive pruning, thus optimizing efficiency without sacrificing accuracy. 

\begin{figure}[H]
  \centering
  \includegraphics[width=\linewidth]{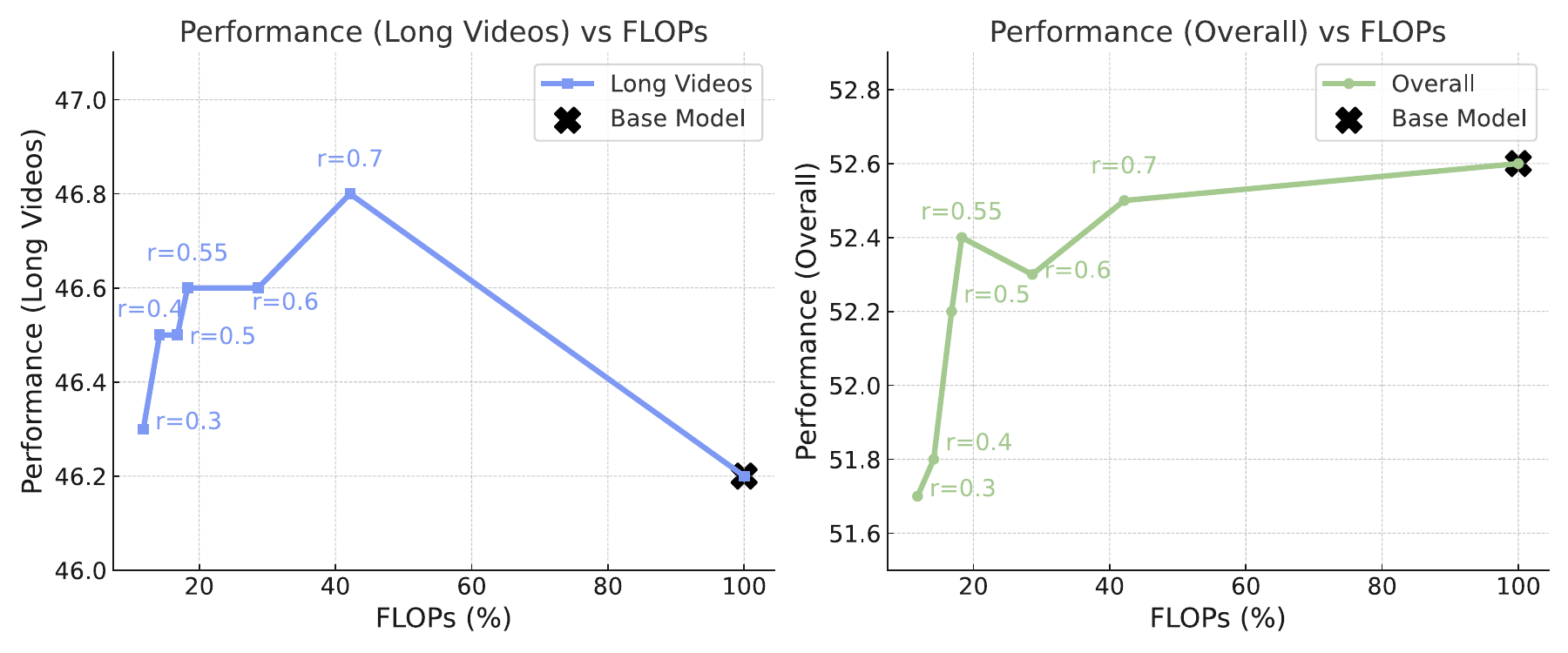}
  \caption{Ablation of different reduction scaling factor $r$ with LongVA-7B on VideoMME, evaluating both overall and long video subset performance.}
  \label{fig:r}
\end{figure}

\noindent\textbf{Selection of Scaling Factor $r$.}
Our ablation study on $r$ shows a trade-off between efficiency and accuracy. Lower values like $r$ = 0.3 or 0.4 overly prune tokens, while higher values like $r$ = 0.7 preserve accuracy but provide limited computational savings. As shown in Figure \ref{fig:r}, we select r = 0.55 as the optimal setting, as it achieves a strong balance, significantly reducing FLOPs while maintaining stable overall performance. Results on long videos further confirm METok’s effectiveness, as compressed versions outperform the base model, suggesting excessive visual tokens may hinder VLMs in long-duration videos.

\begin{table}[!htbp]
    
    \centering
    \begin{adjustbox}{width=\linewidth,center}
    \renewcommand{\arraystretch}{1}
    \setlength{\tabcolsep}{1.5mm}
    \begin{tabular}{lccc}
        \toprule
        \multirow{1}{*}{\textbf{Pruning Strategy}} & \textbf{$L_H=[l_1,l_2,l_3]$} &\textbf{FLOPs} &  \textbf{MLVU} \\
        
        \midrule
         \multirow{1}{*}{Base Model} & without pruning & 100\% & 64.7 \\
        \midrule
        \multirow{3}{*}{Shallow Pruning} & 2,5,10  &  15.5\% & 53.0  \\
        & 3,5,10 & 16.1\%& 53.1\\
        & 3,6,12 & 18.6\% & 55.1 \\
        \midrule
        \multirow{3}{*}{Deep Pruning}& 15,23,26& 46.5\% & 65.1\\
        & 17,23,27& 48.5\% & 65.0\\
        & 19,24,28 & 50.7\% & 65.0\\
        \midrule
        \multirow{2}{*}{Balanced Pruning (Ours)} & 3,10,18 & 26.7\% & 64.5\\
        & \textbf{3,10,19} &  \textbf{27.7\%}&  \textbf{64.7} \\
        \bottomrule
        \end{tabular}
    \end{adjustbox}
    \caption{Ablation of different pruning strategies on LLaVA-OneVision-7B. }
    \vspace{-5pt}
    \label{tab:ablation_layers}
\end{table}

\subsection{Qualitative Results} 
We present qualitative results in Figure~\ref{fig:understanding} on more challenging multi-choice VideoQA tasks in VideoMME and MLVU, showcasing METok’s effectiveness in long-video scenarios. Specifically, our approach accurately performs event segmentation and action counting, capturing key transitions while filtering redundant frames. On the object reasoning task, METok provides stronger multimodal alignment, enabling the model to discern emotional and contextual nuances in human interactions. These observations confirm METok’s stable video-language understanding capabilities. See Appendix~\ref{sec:qual} for more qualitative examples.

\section{Conclusion}
In this paper, we introduced METok, a training-free multi-stage token compression framework that significantly enhances VLLM efficiency for long-video understanding. By integrating event-aware segmentation, hierarchical token pruning, and KV Cache optimization, METok substantially reduces computational costs while preserving accuracy. Our extensive experiments on LLaVA-OneVision-(0.5B, 7B), LongVA-7B and VILA-1.5-13B show that METok achieves up to 80.6\% FLOPs reduction and 93.5\% KV cache savings with comparable performance. METok generalizes well to different model sizes and architectures, while also improving memory consumption for the tested VLLMs. Overall, METok offers a unified and efficient solution that focuses on maintaining long-range temporal and semantic coherence, making it particularly well-suited for long-form video tasks.


\section*{Limitations}
While recent progress has introduced vision large language models (VLLMs) exceeding 13B parameters, our current evaluation is limited to models up to 13B (e.g., VILA-1.5-13B), due to computational resource constraints. These larger models often require substantial GPU memory and infrastructure, which are beyond our reach at this stage. Nonetheless, we conduct comprehensive ablation studies across multiple open-source VLLMs and benchmarks, and the consistent improvements validate the general effectiveness of METok. We leave the evaluation on larger models to future work as resources become available.
\section*{Acknowledgements}
The authors gratefully acknowledge the scientific support and HPC resources provided by the Erlangen National High Performance Computing Center (NHR@FAU) of the Friedrich-Alexander-Universität Erlangen-Nürnberg (FAU) under the NHR project b273dd. NHR funding is provided by federal and Bavarian state authorities. NHR@FAU hardware is partially funded by the German Research Foundation (DFG) – 440719683. Additionally, this work also benefited from the scientific support and HPC resources provided by The Hessian Center for Artificial Intelligence (hessian.AI), as well as the
the Federal Ministry for Research, Technology and Space (BMFTR) project "XEI: Extremely Efficient Inference for Large Context Length" (XEI), project identification number 01IS24079B. This paper is also supported by the DAAD programme Konrad Zuse Schools of Excellence in Artificial Intelligence, sponsored by the Federal Ministry of Research, Technology and Space.
\bibliography{latex/custom}

\clearpage
\appendix

\section{Datasets Details}
\label{sec:dataset}
We conduct comprehensive experiments comparing METok's performance with other training-free token compression methods on these video understanding benchmarks:
\begin{itemize}[nosep]
\item MVBench~\cite{mvbench}: MVBench is a diagnostic benchmark for evaluating the temporal understanding abilities of multimodal large language models. It consists of 20 challenging tasks derived from existing static image benchmarks, transformed into video-based versions that require dynamic perception and reasoning. Each task uses a multiple-choice QA format based on curated YouTube video clips, with durations ranging from short to medium length. We conduct the zero-shot evaluation on the test set.
\item  EgoSchema~\cite{egoschema}: EgoSchema is a long-form video QA benchmark built on egocentric videos from the Ego4D dataset. It includes over 5,000 multiple-choice questions requiring reasoning over 3-minute video clips that capture real-world daily activities. Each question is paired with five answer options, and a subset of 500 questions includes human-annotated ground-truth labels. Following the standard evaluation setup, we report zero-shot performance on the test set.
\item MLVU~\cite{zhou2024mlvu}: MLVU is a diagnostic benchmark designed to test different aspects of long video comprehension, including event identification, temporal reasoning, and narrative understanding. It features thousands of video clips ranging from 30 seconds to several minutes, with questions that span low-level recognition to high-level inference. We adopt the zero-shot setting and follow the official protocol on its dev set for evaluation.
\item Video-MME~\cite{videomme}: Video-MME is a large-scale benchmark designed to comprehensively assess the video understanding capability of VLLMs. It contains 900 videos spanning 254 hours in total, with durations ranging from 11 seconds to over 60 minutes. The dataset covers 6 high-level categories and 30 subcategories, and includes manually annotated multiple-choice questions that require spatial-temporal reasoning and multi-event analysis. Our evaluation follows the official test set in a zero-shot manner.
\end{itemize}

\section{Pseudocode}
We show the pseudo code of Event-aware Token Reduction during vision encoding stage in Algorithm \ref{alg:metok-vision}.

{
\begin{footnotesize}
\begin{algorithm}[!htbp]
\caption{METok - Vision Encoding}
\label{alg:metok-vision}
\KwInput{Frame-level visual embedding $ \{v_1, \dots, v_T\}$; text embedding $t$}
\KwParam{Number of events $k$; key event ratio $\alpha$; key frame ratio $\beta$; pooling strides $s_1$, $s_2$}
\KwOutput{Retained visual tokens $V_{\text{r}}$}

\vspace{4pt}
\Comment{Event temporal segmentation}
\vspace{2pt}
$S_i \gets \cos(v_i, v_{i+1}), \quad \forall i \in [1, T{-}1]$\;
$E \gets \text{Split}(V)$ at $(k{-}1)$ lowest $S_i$ positions\;

\vspace{4pt}
\Comment{Key semantic identification}
\vspace{2pt}
\For{$E_j \in E$}{
    \For{$v_i \in E_j$}{
        $S_{v_i,t} \gets \cos(v_i, t)$\;
    }
    $r_j \gets \max(S_{v_i,t})$\;
    $F_j^{\text{key}} \gets$ top-$\beta$ frames in $E_j$ by $S_{v_i,t}$\;
    $F_j^{\text{non-key}} \gets E_j \setminus F_j^{\text{key}}$\;
}
$E_{\text{key}} \gets$ top-$\lceil \alpha k \rceil$ events by $r_j$\;
$E_{\text{non-key}} \gets E \setminus E_{\text{key}}$\;

\vspace{4pt}
\Comment{Adaptive pooling strategy}
\vspace{2pt}
$V_{\text{r}} \gets [\,]$\;
\For{$E_j \in E$}{
  \If{$E_j \in E_\text{key}$}{
    \For{$v_i \in E_j$}{
      \uIf{$v_i \in F_j^{\text{key}}$}{
        $v_i^{\text{p}} \gets \text{pool2d}(v_i, s_1)$\;
      }
      \Else{
        $v_i^{\text{p}} \gets \text{pool2d}(v_i, s_2)$\;
      }
      Append $v_i^{\text{p}}$ to $V_{\text{r}}$\;
    }
  }
  \Else{
    \For{$v_i \in E_j$}{
      \uIf{$v_i \in F_j^{\text{key}}$}{
        $v_i^{\text{p}} \gets \text{pool2d}(v_i, s_1/\alpha)$\;
      }
      \Else{
        $v_i^{\text{p}} \gets \text{pool2d}(v_i, s_2/\alpha)$\;
      }
      Append $v_i^{\text{p}}$ to $V_{\text{r}}$\;
    }
  }
}
\Return{$V_{\text{r}}$}
\end{algorithm}
\end{footnotesize}
}
\section{Implementation Details}
\label{sec:implementation}
LLaVA-OneVision-(0.5B, 7B) and LongVA-7B are all built on top of Qwen. LLaVA-OneVision-7B and LongVA-7B use 28 transformer layers, while LLaVA-OneVision-0.5B has 24 layers. VILA-1.5-13B is based on LLaMA2~\cite{touvron2023llama} and consists of 40 transformer layers.

We use a single set of hyperparameters for each base model across all benchmarks without task-specific tuning. For both LLaVA-OneVision-(0.5B, 7B) and LongVA-7B, we use $(s_1, s_2) = (2, 3)$, $L_H = [3, 10, 19]$, and $\alpha = 0.5$ as shared hyperparameters. The remaining parameters vary by model: for LLaVA-OneVision-(0.5B, 7B), we set $k=5$, $\beta=0.4$, and $r=0.76$; for LongVA-7B, we set $k=13$, $\beta=0.45$, and $r=0.55$. For VILA-1.5-13B, we use the setting of $(s_1, s_2) = (2, 3)$, $L_H = [13, 24, 34]$, $\alpha = 0.5$, $k=3$, $\beta=0.4$, and $r=0.65$. 


All experiments are conducted on NVIDIA A100 80 GPUs. To evaluate efficiency, we report FLOPs, Prefill Time, and KV Cache memory averaged across multiple video understanding benchmarks, including MLVU, VideoMME, MVBench, and EgoSchema.

\section{More Details of Experiments}
\subsection{Scalability to More Frames.}
To assess METok’s ability to handle longer videos under limited context, we conduct experiments on VILA-1.5-13B, which is based on LLaMA2 and supports a maximum context length of 4096 tokens. Under the default setting, VILA-1.5 accommodates only 16 frames without truncating visual or textual inputs. As shown in Table~\ref{tab:frame_scaling}, by integrating METok, we have tested to scale the number of input frames from 16 to 64 within the same context budget.

Notably, as the number of frames increases, METok not only maintains performance but also brings consistent gains, especially on the MLVU and VideoMME benchmarks. For instance, at 64 frames, METok improves the MLVU score from 50.4 to 53.6, and VideoMME from 49.5 to 50.7. These results indicate that METok consistently achieves strong accuracy across different frame sampling densities and effectively compresses less informative tokens while preserving task-relevant visual semantics, allowing the model to benefit from richer temporal context. 

Importantly, the tested frame counts (up to 64) do not yet represent the upper bound of METok’s capability, and further improvements are expected with higher frame inputs. This demonstrates METok’s strong scalability under strict context constraints and highlights its potential for long-form video understanding.

Below, we further compare different token reduction methods with VILA-1.5-13B under the 32-frame inputs. As shown in Table~\ref{tab:tokenreduction-32f}, METok achieves the lowest FLOPs while outperforming others in MVBench, MLVU, and VideoMME benchmarks. Notably, VILA-1.5-13B has a native maximum context length of 4096, which limits it to processing only 16 frames without any token reduction. Within this controlled setting, METok delivers better trade-off between accuracy and computation, indicating that its multi-stage, event-aware compression preserves task-relevant cues while minimizing redundant computation.
\begin{table}[!htbp]
    \centering
    \small
    \begin{adjustbox}{width=\linewidth,center}
    \renewcommand{\arraystretch}{1}
    \setlength{\tabcolsep}{1.5mm}
    \begin{tabular}{lcccc}
        \toprule
        \textbf{Method} & \textbf{\#Frames} & \textbf{MVBench} & \textbf{MLVU} & \textbf{VideoMME} \\
        \midrule
        Base Model & 16 & 50.9 & 50.4 & 49.5 \\
        w/METok & 16 & 50.5 & 50.5 & 49.5 \\
        w/METok & 32 & 50.9 & 52.8 & 49.9 \\
        w/METok & 48 & 51.2 & 53.4 & 50.2 \\
        \rowcolor{YellowGreen!20} w/METok & 64 & 50.7 & 53.6 & 50.7 \\
        \bottomrule
    \end{tabular}
    \end{adjustbox}
    \caption{Evaluation of METok under increasing frame numbers on VILA-1.5-13B.}
    \label{tab:frame_scaling}
\end{table}
\vspace{-3pt}
\begin{table}[!htbp]
  \centering
  \begin{adjustbox}{width=\linewidth,center}
  \setlength{\tabcolsep}{1.5mm}
  \renewcommand{\arraystretch}{1.15}
  \begin{tabular}{lccccc}
    \toprule
    \textbf{Method} &  \textbf{FLOPs (TB)} & \textbf{MVBench} & \textbf{MLVU} & \textbf{VideoMME} \\
    \midrule
    w/FastV         & 78.0 & 50.4 & 52.0 & 48.9 \\
    w/VisionZip      & 56.7 & 48.9 & 50.3 & 47.9 \\
    w/DivPrune      & 37.2 & \textbf{50.9} & 52.5 & 49.4 \\
    \rowcolor{YellowGreen!20} w/METok  & \textbf{36.5} & \textbf{50.9} & \textbf{52.8}& \textbf{49.9} \\
    \bottomrule
  \end{tabular}
  \end{adjustbox}
  \caption{Comparison of token-reduction methods with VILA-1.5-13B under 32-frame inputs.}
  \label{tab:tokenreduction-32f}
\end{table}

\begin{table*}[!t]
  \centering
  \small
  \setlength{\tabcolsep}{6pt}
  {\renewcommand{\arraystretch}{1.1}
  \begin{tabular}{lccccccccc}
    \toprule
    Method & FLOPs (TB) & AR & NQA & TR & PQA & AO & AC & ER & Overall \\
    \midrule
    VILA-1.5-13B      & 84.1 & \textbf{52.5} & 53.0 & 78.3 & 57.1 & 35.9 & \textbf{23.6} & \textbf{52.0} & 50.4 \\
    w/FastV           & 49.8 & 52.0 & 52.4 & 78.7 & 56.4 & 35.0 & 22.0 & 50.0 & 49.5 \\
    w/VisionZip       & 29.1 & 46.5 & 51.3 & 72.2 & 50.7 & \textbf{36.9} & 22.0 & 50.0 & 47.1 \\
    w/DivPrune        & 20.3 & 51.4 & 54.6 & 78.7 & 57.2 & 35.8 & 22.2 & 51.9 & 50.3 \\
    \rowcolor{YellowGreen!20} w/METok (Ours)    & \textbf{19.6} & 50.6 & \textbf{55.3} & \textbf{78.8} & \textbf{57.8} & \textbf{36.9} & 22.5 & 51.3 & \textbf{50.5} \\
    \bottomrule
  \end{tabular}
  }
  \caption{Results on MLVU subsets with VILA-1.5-13B.}
  \label{tab:metok-results}
\end{table*}
\vspace{-5pt}
\subsection{Key Event and Frame Retention Ratios.}
We conduct a joint ablation on the key event ratio $\alpha$ and key frame ratio $\beta$ to study their impact on both efficiency and performance with LLaVA-OneVision-7B. As shown in Table~\ref{tab:event_frame_ablation}, lower values of $\alpha$ and $\beta$ reduce FLOPs more aggressively but lead to noticeable drops in accuracy, particularly on VideoMME. In contrast, increasing $\alpha$ beyond 0.5 provides marginal performance gains while incurring higher computation cost.

The best trade-off is achieved at $\alpha = 0.5$ and $\beta = 0.4$, which offers a 70\% FLOPs reduction compared to the base model and consistently outperforms other settings. These results suggest that METok is stable across a reasonable range of hyperparameters, and its performance does not rely on fine-tuned thresholds. Notably, even with $\alpha = 0.3$ and $\beta= 0.2$, METok still maintains solid accuracy, highlighting the flexibility of our key visual-text semantic identification design. 

\begin{table}[!htbp]
\centering
\begin{adjustbox}{width=\linewidth,center}
\renewcommand{\arraystretch}{1.1}
\setlength{\tabcolsep}{1.5mm}
\begin{tabular}{ccccc}
\toprule
\textbf{$\alpha$} & \textbf{$\beta$} & \textbf{FLOPs (TB)} & \textbf{Egoschema} & \textbf{VideoMME} \\
\midrule
\rowcolor{gray!10}
\multicolumn{2}{l}{\textit{Base Model (32 frames)}} & 71.4 & 43.5 & 58.2 \\
\midrule
0.3 & 0.2 & 14.6 & 42.8 & 57.3 \\
0.3 & 0.4 & 17.9 & 43.3 & 57.6 \\
0.5 & 0.2 & 17.2 & 43.7 & 58.2 \\
\rowcolor{YellowGreen!20}
0.5 & 0.4 & 21.8 & 44.4 & 58.4 \\
0.7 & 0.2 & 34.5 & 44.3 & 58.2 \\
0.7 & 0.4 & 43.3 & 44.4 & 58.7 \\
\bottomrule
\end{tabular}
\end{adjustbox}
\caption{Ablation study on key event ratio $\alpha$ and key frame ratio $\beta$ using METok on LLaVA-OneVision-7B with 32-frame input.}
\label{tab:event_frame_ablation}
\end{table}

\subsection{Performance across MLVU subsets}
To provide a more comprehensive evaluation, we conducted a detailed analysis across all seven MLVU subcategories. As shown in Table~\ref{tab:metok-results}, METok achieves the highest or competitive accuracy in most task types and consistently outperforms all baselines in overall accuracy and FLOPs reduction (from 84.1 to 19.6 TB). It particularly excels in long-context reasoning tasks such as NQA and PQA, where its token compression effectively preserves key visual-text semantics over time. These results not only demonstrate the task-level robustness and generality of METok, but also highlight its strength in long-form video question answering.

\section{Additional Qualitative Examples}
\label{sec:qual}
\begin{figure}[!htbp]
  \centering
  \includegraphics[width=\linewidth]{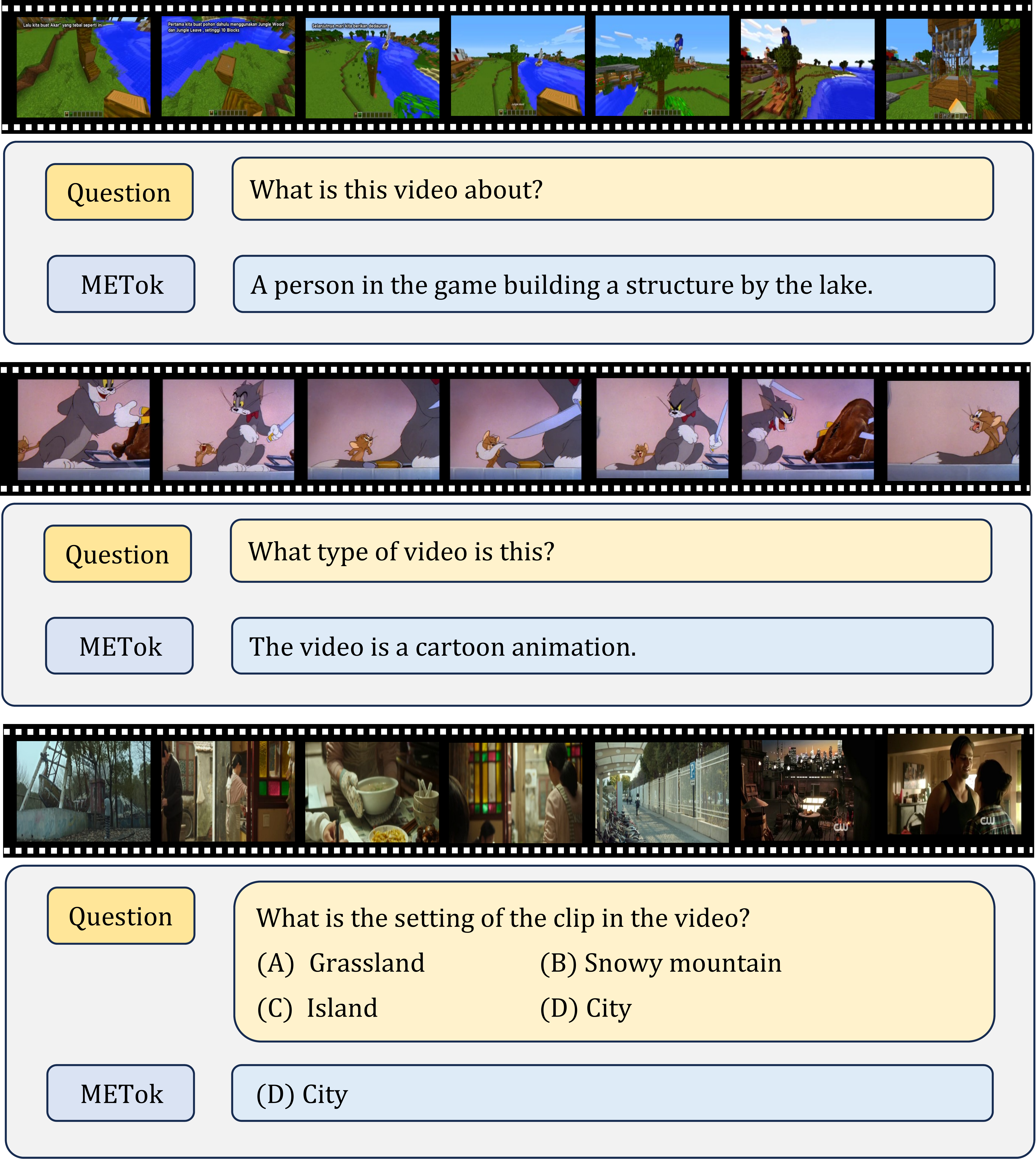}
  \caption{More video understanding example of our proposed method METok with VILA-1.5-13B.}
  \label{fig:qual2}
\end{figure}

\begin{figure}[h]
  \centering
  \includegraphics[width=\linewidth]{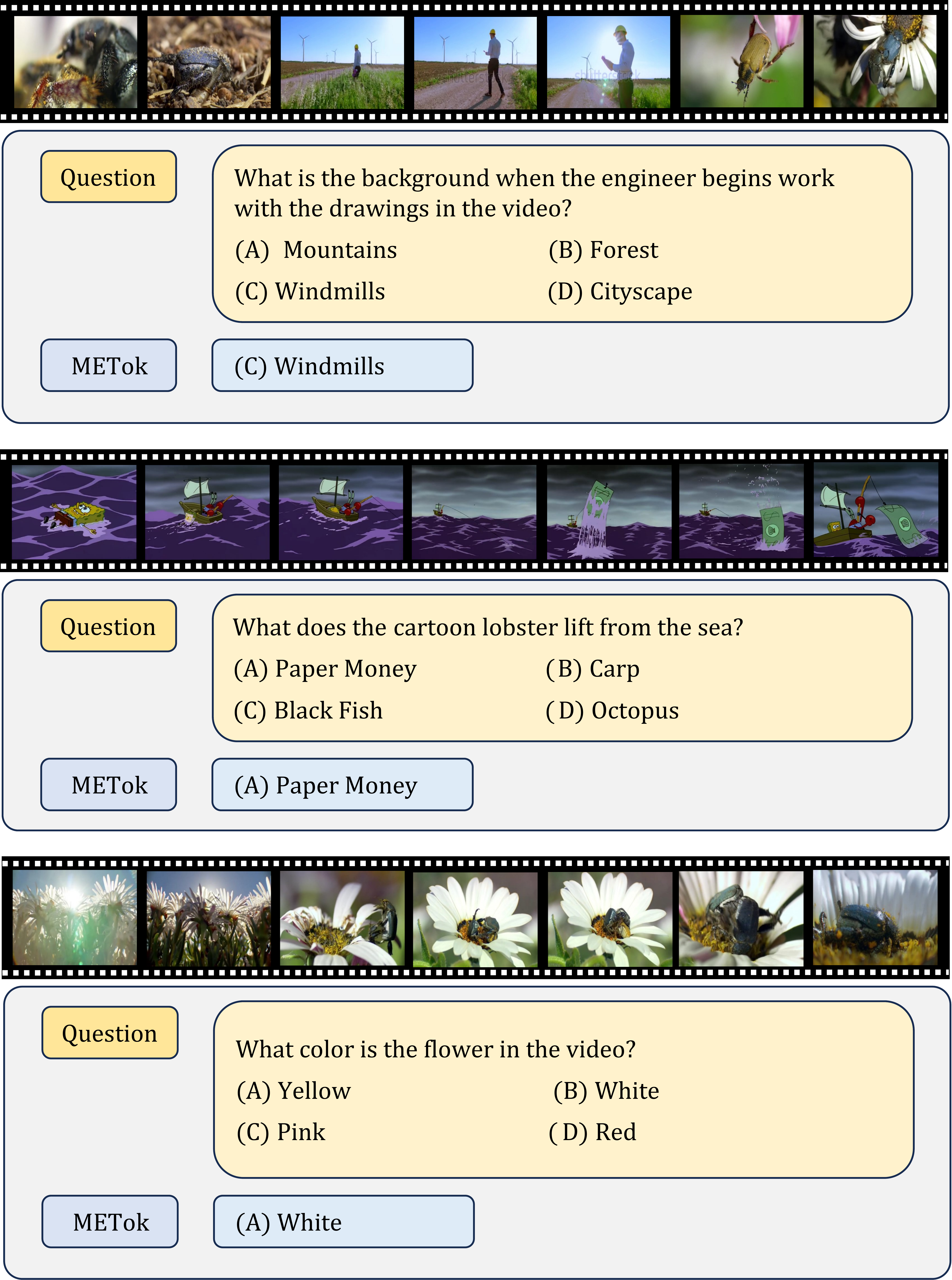}
  \caption{More video question answering example from the MLVU~\cite{zhou2024mlvu} of our proposed method METok with LongVA-7B.}
  \label{fig:qual1}
\end{figure}

We provide additional video question answering examples from video understanding benchmarks like MLVU, as shown in Figure~\ref{fig:qual2} and Figure~\ref{fig:qual1}. In the background recognition example, METok with LongVA-7B correctly selects “Windmills” as the scene behind the engineer working with drawings, indicating robust scene-level understanding under distractor options. In the object retrieval case from an animated domain, METok accurately identifies that the cartoon lobster lifts paper money, showing resilience to style shifts and the ability to ground fine object semantics. For an attribute recognition query, METok answers white for the flower color, demonstrating precise grounding of low-level visual attributes. Together, these cases span scene context, object semantics, and fine-grained attributes, and they align with our design goal of preserving key visual-text cues while suppressing redundancy in long videos.

These examples further demonstrate the effectiveness of the proposed METok framework in capturing key visual-semantic cues from long videos. METok consistently produces precise and contextually grounded answers, successfully identifying fine-grained actions, object interactions, and temporal dependencies. By focusing on the relevant visual segments and filtering out redundant frames, METok generates accurate and coherent responses.

\end{document}